\documentclass[11pt]{article}
\usepackage[margin=1in]{geometry}
\usepackage{caption}
\usepackage{graphicx}
\usepackage{xcolor}
\definecolor{RevisionBlue}{RGB}{15,67,155}

\usepackage{url}
\usepackage{cite}
\usepackage[hidelinks]{hyperref}
\usepackage{amsmath}
\usepackage{amsfonts}

\title{XANE(3): An $E(3)$-Equivariant Graph Neural Network for Accurate Prediction of XANES Spectra from Atomic Structures}

\author{
Vitor F. Grizzi, Luke N. Pretzie, Jiayi Xu, and Cong Liu\thanks{Corresponding author. Email: congliu@anl.gov} \\[0.5em]
Chemical Sciences and Engineering Division \\
Argonne National Laboratory, Lemont, IL 60439, USA
}

\begin{document}

\maketitle
\begin{abstract}
Predicting X-ray absorption near-edge structure (XANES) across chemically and structurally diverse materials requires accurate spectral reconstruction alongside sensitivity to local atomic environments. We present XANE(3), an $E(3)$-equivariant graph neural network that combines absorber-conditioned representations, an element-routed mixture-of-experts readout, and a derivative-aware training objective. Trained on 617,203 Materials Project FEFF9 spectra spanning 88 elements and K, L$_2$, and L$_3$ edges, XANE(3) achieves a held-out intensity MSE of $2.38\times10^{-3}$ and an absolute energy-origin MAE of 0.14~eV. Systematic ablations show that higher-order tensorial features substantially improve pointwise intensity accuracy at matched parameter counts, whereas first- and second-derivative penalties improve slope and curvature fidelity at a modest cost in intensity MSE. On an independent FDMNES iron-oxide benchmark spanning bulk crystals, undercoordinated surfaces, and perturbed geometries, readout-only adaptation with a frozen encoder achieves a 7.3-fold lower intensity MSE than an optimal site-agnostic baseline, confirming reusable site-specific information, while compact target-trained models yield the lowest aggregate errors. These results establish XANE(3) as a general structure-to-spectrum surrogate and clarify when bulk-pretrained representations transfer effectively to heterogeneous materials.
\end{abstract}

\section{Introduction}

X-ray absorption spectroscopy (XAS) is a widely used, element-specific probe of the local coordination environment and electronic structure of molecules and materials. XAS spectra are typically divided into two regions: X-ray absorption near-edge structure (XANES), which arises primarily from transitions of core-level electrons to unoccupied bound or quasi-bound states, and extended X-ray absorption fine structure (EXAFS), which is dominated by transitions to continuum states and reflects local atomic arrangement. Because XANES encodes oxidation state, coordination number, local symmetry, and multiple-scattering fingerprints, it is particularly valuable for inferring local structure when crystallography is inconclusive, when materials are disordered, or when systems evolve under operating conditions~\cite{zhu2021k, harper2023modelling, xu2024recent}. In heterogeneous catalysis, for example, XANES is widely used to track oxidation-state changes and dynamic restructuring of active sites under reaction conditions, including variations in temperature, pressure, and chemical potential~\cite{timoshenko2020situ, shi2021dynamics, xu2024theoretical, khurana2024unveiling, patel2022integrated, meng2024multicode}. Even as an ensemble-averaged probe, its element selectivity provides strong constraints on plausible mechanistic hypotheses~\cite{xu2023understanding}.

Historically, XANES simulations have relied on first-principles methods implemented in codes such as ORCA, FEFF, FDMNES, and OpenMolcas~\cite{ORCA, rehr2010parameter, bunuau2009self, guda2015optimized, li2023openmolcas}, drawing on time-dependent density functional theory (TDDFT), real-space multiple scattering, finite-difference methods, and multireference wavefunction theory~\cite{rehr2009ab, rehr2000theoretical, xu2024recent2}. Despite their accuracy, these calculations are computationally demanding because the electronic structure (often including core-hole effects and unoccupied states) must be resolved separately for each atomic configuration. This bottleneck is amplified in heterogeneous materials such as nanoparticle catalysts, liquids, and amorphous phases, where realistic theoretical spectra require sampling broad distributions of local environments and many snapshots from molecular dynamics trajectories~\cite{smith2017soft}. The burden extends equally to quantitative spectral interpretation, which typically relies on repeated forward simulations to test candidate structures involving defects, surfaces, adsorbates, or dynamic ensembles against experiment, making high-throughput screening and iterative structure refinement costly. While recent agentic frameworks such as ChemGraph-XANES have streamlined simulation setup for physics-based spectroscopy campaigns~\cite{grizzi2026chemgraph}, the underlying cost of repeated first-principles calculations remains a fundamental bottleneck.

These limitations have motivated growing interest in machine learning (ML) for both forward spectral prediction and inverse structural inference~\cite{kotobi2023integrating, torrisi2020random, guda2021understanding, na2025interpretable, carbone2020machine, rankine2022accurate, kulaev2026deepfit, zhan2025graph}. Inverse models extract chemically meaningful descriptors—such as coordination number, bond lengths, or oxidation state—directly from spectra or multimodal probes~\cite{torrisi2020random, guda2021understanding, na2025interpretable}, whereas forward models approximate the structure-to-spectrum mapping from computed or experimental databases~\cite{carbone2020machine, rankine2022accurate, zhan2025graph}. Recent forward-prediction frameworks have expanded this paradigm across elements and simulation fidelities: OmniXAS combines learned structural representations with multi-element and cross-fidelity transfer learning~\cite{kharel2025omnixas}, while Lin et al. developed a crystal graph model spanning 48 elements and calibrated it with limited experimental data~\cite{lin2026experimentally}. Unlike conventional scalar-property regression, however, spectral prediction targets structured, high-dimensional functions subject to alignment and normalization ambiguities. Consequently, chemical coverage, pointwise reconstruction accuracy, spectral-shape fidelity, and performance after domain adaptation are distinct criteria that require separate evaluation.

Modern ML for atomistic prediction can be viewed along two coupled design axes: how atomic structure is represented, and what physical symmetries are built into the model as inductive biases~\cite{grizzi2026nugraph2}. A common baseline pairs handcrafted local descriptors—such as radial distribution encodings, atom-centered symmetry functions, or cumulative distribution features—with multilayer perceptrons (MLPs) or random forests~\cite{chen2024robust}. Although data-efficient, these approaches are limited by fixed cutoff radii and non-adaptive functional forms. Graph neural networks (GNNs) overcome these constraints by operating directly on atomic graphs and jointly learning structural representations and target mappings through message passing~\cite{reiser2022graph, murg2025enhanced, grizzi2026active, grizzi2026alf}. Equivariant GNNs provide an even stronger inductive bias for three-dimensional atomic structures: architectures equivariant under $E(3)$ or $SE(3)$ enforce prescribed transformations of latent features under rotations and translations (and reflections for $E(3)$). Embedding these transformation laws directly (e.g., using spherical harmonics and tensor products in frameworks such as e3nn~\cite{geiger2022e3nn}) improves data efficiency, generalization, and robustness even when the final target, such as total energy or scalar XANES intensity, is globally invariant~\cite{duval2023hitchhiker}.

Transfer beyond bulk-crystal environments poses a further challenge for materials spectroscopy. At iron-oxide surfaces, for example, undercoordination and local geometric distortions can substantially alter Fe K-edge XANES~\cite{chen2002fe2o3}, and resolving inequivalent metal sites is directly relevant to catalysis because distinct surface sites can participate differently in individual reaction steps~\cite{liu2023constructing}. A central question is whether representations learned from bulk-crystal spectra retain the geometric information needed to predict spectra of undercoordinated and structurally perturbed sites. Adapting only the spectral readout while keeping the encoder frozen provides a direct test of the information preserved in the pretrained representation, while full fine-tuning and compact models trained directly on the target domain provide complementary benchmarks for attainable accuracy. Such comparisons also require careful accounting for changes in simulation approximations, energy references, and preprocessing that accompany a shift in structural domain.

To address these challenges, we developed XANE(3), a XANES prediction framework built on an $E(3)$-equivariant GNN backbone. The architecture combines equivariant message passing, tensor-product interactions, and spherical harmonic features with custom equivariant layer normalization and adaptive gated residual connections to preserve higher-order geometric information. Absorber-query attention conditions the pooled structural context on the selected absorber, an element-routed mixture-of-experts readout predicts normalized intensities on a relative energy grid, and a detached auxiliary head predicts the absolute energy origin. We first train and evaluate XANE(3) on 617,203 Materials Project FEFF9 bulk-crystal spectra spanning 88 elements and K, L$_2$, and L$_3$ edges, systematically ablating architectural components and derivative-aware shape penalties. We then assess transfer to an independent FDMNES Fe K-edge dataset of bulk and surface iron oxides (Fe$_2$O$_3$, Fe$_3$O$_4$, and FeO), which are of broad relevance to catalysis, geochemistry, and energy materials~\cite{piquer2014fe, huang2024electrosynthesis, zhu2021k}. Using this dataset, we evaluate undercoordinated surface sites and perturbed geometries under readout-only adaptation, full fine-tuning, and compact training from scratch.

\section{Methods}
XANE(3) maps an atomic structure, a selected absorber or absorber-site set, and a core-level label to a normalized XANES intensity vector. An equivariant graph encoder represents the atomic geometry, absorber-conditioned pooling constructs invariant readout features, and a spectral head predicts intensities on a fixed relative-energy grid. A separate auxiliary head predicts the stored energy origin used to reconstruct the absolute energy axis. The spectral shape and energy origin are distinct prediction targets, defined by the preprocessing conventions below. The learned mapping applies to the structural and simulation conditions represented in the reference data.

\subsection{Dataset and preprocessing}
\subsubsection{Dataset generation}
The Materials Project pretraining corpus comprised FEFF9 XANES spectra and their associated bulk crystal structures~\cite{mathew2018high,chen2021database}. After preprocessing, the corpus contained 617,203 structure--spectrum pairs across 88 absorber elements, including 466,117 K-edge, 75,519 L$_2$-edge, and 75,567 L$_3$-edge spectra. The pretraining split contained 537,203 spectra, of which 5,105 (0.95\%) were Fe K-edge spectra, while the validation and test splits each contained 40,000 spectra. Splits were stratified at the individual-spectrum level by absorber element and core level and, for sufficiently populated strata, by atom count and energy (the reference edge energy when available, otherwise the first sampled energy).

To test transfer to undercoordinated and perturbed Fe environments, we assembled an iron-oxide target dataset of 5,919 Fe K-edge spectra from 539 distinct structures spanning bulk crystals, surface slabs, and perturbed geometries. For the slab calculations, we used Materials Project bulk structures of $\alpha$-, $\beta$-, and $\gamma$-Fe$_2$O$_3$, Fe$_3$O$_4$, and FeO~\cite{Jain,ong2015materials}. Six-layer slabs with (001), (012), (104), and (110) facets were generated with pymatgen~\cite{ong2013pymatgen,Tran} (median inter-slab vacuum gap of 29.9~\AA{}, range 17.3--30.8~\AA{}) and relaxed using the OMat task of the UMA machine-learned interatomic potential~\cite{Wood,BarrosoLuque} to a 0.05~eV/\AA{} force threshold. For each facet, 10 perturbed configurations were generated via Gaussian displacements ($\sigma = 0.01$--$0.1$~\AA{} for atomic positions and $0.02$--$0.1$~\AA{} for lattice vectors) to sample off-equilibrium local distortions rather than a thermal ensemble.

Fe K-edge XANES simulations for the iron-oxide structures were performed using the FDMNES code~\cite{Joly}, with separate calculations conducted for each symmetrically distinct Fe absorber. Self-consistent-field calculations employed the Green's-function approach within the spherical-wave approximation, including quadrupole transitions alongside the dominant dipole transitions. The calculated energy range extended from $-55$ to $+150$~eV relative to the Fermi energy, using step sizes of 1.0~eV from $-55$ to $-10$~eV, 0.01~eV from $-10$ to $+5$~eV, and 0.1~eV from $+5$ to $+150$~eV. Theoretical spectra were convolved to account for core-hole lifetime broadening and experimental resolution, and the resulting spectra were used in the iron-oxide transfer benchmark (Sec.~\ref{sec:iron-oxide-transfer-methods}).

\subsubsection{Spectral Preprocessing}
The Materials Project records were screened to retain spectra with valid structural and absorber metadata and at least three unique finite energy-intensity points. Nonfinite energy or intensity values were removed, and spectra were sorted by increasing energy. For each spectrum, we subtracted the lowest sampled energy from all energy values so that the spectrum began at 0 eV. To reduce variations in absolute spectral amplitude while preserving spectral shape, each spectrum was divided by its 99th-percentile intensity before interpolation. Spectra containing negative absorption intensities, a nonpositive normalization factor, nonfinite normalized values, or a normalized maximum intensity greater than 2 were excluded. The baseline used a fixed model grid of 100 uniformly spaced points from 0 to 55~eV. For the iron-oxide benchmark, we first interpolated each FDMNES spectrum onto 100 equally spaced points from $-5$ to $+50$~eV relative to the stored Fe K-edge reference of 7112~eV. We then divided the interpolated intensities by their 99th percentile and shifted the grid to the model's 0--55~eV axis. 

\subsubsection{Graph Preprocessing}
For each simulation, we extracted (i) the atomic structure, including lattice parameters and fractional coordinates, (ii) the absorber identity and site index, (iii) the computed XANES spectrum, and (iv) the edge energy $E_0$ from the simulation output when available, storing the processed entries in an ASE SQLite database~\cite{hjorth2017atomic}. Each structure was converted into a periodic graph in which nodes represent atoms and edges connect neighbors within a radial cutoff $r_{\max} = 5.0$~\AA{}, with displacement vectors accounting for periodic image shifts. Each graph corresponded to one reference spectrum, with a binary node mask identifying the associated absorber site or sites.

The primary target for each graph was the normalized intensity vector on the model energy grid. For the pretrained baseline, we also retained the absolute energy of the first sampled spectrum point as an auxiliary scalar target, standardized separately for each absorber element and core level.

\subsection{Model Architecture}
\begin{figure}[!htb]
    \centering
    \includegraphics[width=\linewidth]{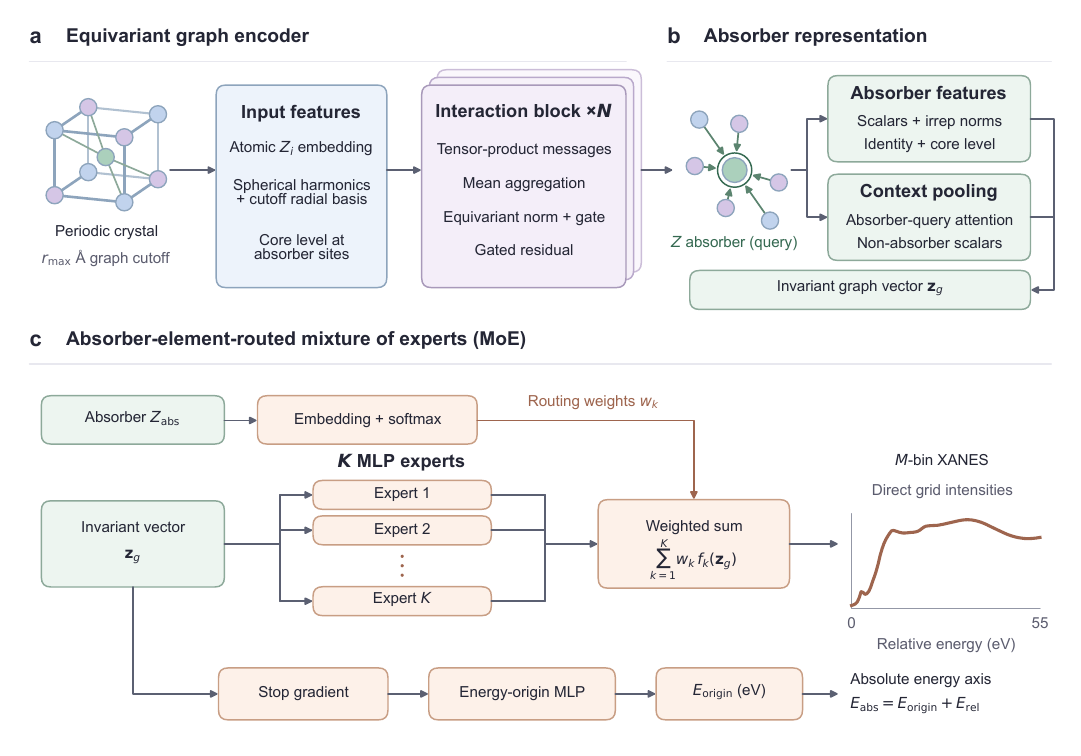}
    \caption{XANE(3) with the direct spectral readout. (a) The periodic atomic graph uses a neighbor cutoff of $r_{\max}=5$~\AA{} and three equivariant interaction blocks. (b) The arrows from the six closest graph neighbors of the highlighted Fe absorber illustrate absorber-query attention. The model pools scalar features from all non-absorber sites and combines the result with the absorber scalars, higher-order irrep norms, absorber identity, and core level to form the invariant graph vector $\mathbf{z}_g$. (c) All eight readout experts receive $\mathbf{z}_g$. The absorber atomic number determines the weights used to sum their bin predictions. The auxiliary head receives the detached graph vector and predicts the stored energy origin for reconstructing the absolute energy axis. The trace shows the predicted Fe K-edge spectrum for the absorber illustrated in (b).}
    \label{fig:overview}
\end{figure}

We model each structure as an atomic graph $G=(V,E)$, where node $i \in V$ is associated with an atomic number $Z_i$ and Cartesian position $\mathbf{r}_i$. Edges are constructed between atoms within a cutoff radius $r_{\max}$. For periodic systems, edge vectors are corrected by the lattice image displacement stored for each graph edge,
\[
\mathbf{x}_{ij} = \mathbf{r}_j - \mathbf{r}_i + \mathbf{s}_{ij}, 
\qquad
d_{ij} = \|\mathbf{x}_{ij}\|,
\]
where $\mathbf{s}_{ij}$ is the Cartesian periodic-image displacement. The model predicts a XANES spectrum from the absorber environment by combining an $E(3)$-equivariant message-passing backbone, absorber-conditioned pooling, spectroscopy core-level conditioning, and an absorber-element-routed spectral readout. Figure~\ref{fig:overview} shows the direct-readout configuration used for the Materials Project baseline. The equivariant encoder can carry directional relationships through its hidden layers and combine them into invariant features. For example, the scalar contraction of two vector features can encode the angle between neighboring bonds, preserving sensitivity to local geometry under global rotation.

\subsubsection{Initial atomic and core-level representation}
Each atom is initialized from its atomic number through a learnable embedding table,
\[
\mathbf{h}_{i,\mathrm{atom}}^{(0)} = \mathrm{Emb}_Z(Z_i),
\]
where $\mathbf{h}_{i,\mathrm{atom}}^{(0)}$ contains only scalar channels. In contrast to one-hot encoding, this embedding allows chemically related elements to acquire related latent representations during training.

For mixed-edge datasets, each graph also carries a spectroscopic core-level label $c_g$, such as K, L$_2$, or L$_3$. This label is mapped to a learned embedding,
\[
\mathbf{e}_{c,g} = \mathrm{Emb}_c(c_g),
\]
which is projected to the scalar node-feature width and added to absorber atoms only:
\[
\mathbf{h}_i^{(0)} =
\begin{cases}
\mathbf{h}_{i,\mathrm{atom}}^{(0)} + \phi_c(\mathbf{e}_{c,g})
& i \in \mathcal{A}_g,\\
\mathbf{h}_{i,\mathrm{atom}}^{(0)}
& i \notin \mathcal{A}_g.
\end{cases}
\]
Here $\mathcal{A}_g$ denotes the set of absorber atoms in graph $g$, and $\phi_c$ is a small MLP. The unconditioned atomic embedding is also retained and later averaged over absorber sites, giving the readout an explicit absorber-identity signal independent of the message-passing state.

\subsubsection{Equivariant interaction block}
\paragraph{Message passing.}
The backbone consists of $L$ interaction blocks. At each layer $\ell$, node features live in the direct-sum representation space
\[
\mathcal{H}
=
m_0 \times 0e
\oplus
m_1 \times 1o
\oplus
m_2 \times 2e,
\qquad
\mathbf{h}_i^{(\ell)} \in \mathcal{H}.
\]
Here $m_0$, $m_1$, and $m_2$ are user-defined multiplicities for the scalar ($l=0$), vector ($l=1$), and rank-2 tensor ($l=2$) channels, respectively. For an edge $(i \to j)$, the message is computed using a tensor product between the source features and the spherical harmonics of the unit edge direction $\hat{\mathbf{x}}_{ij} = \mathbf{x}_{ij}/d_{ij}$,
\[
\mathbf{m}_{ij}^{(\ell)}
=
\mathrm{TP}\!\left(
\mathbf{h}_i^{(\ell)},
Y(\hat{\mathbf{x}}_{ij});
\psi(d_{ij})
\right),
\]
where $Y(\hat{\mathbf{x}}_{ij})$ denotes spherical harmonics up to order $l_{\max}$ and $\psi(d_{ij})$ are tensor-product weights generated from a radial basis expansion of the interatomic distance. The reported models use Bessel radial basis functions with a smooth cosine cutoff. Incoming messages are mean-aggregated over the neighborhood $\mathcal{N}(j)$ of each destination node $j$,
\[
\mathbf{m}_j^{(\ell)} = \frac{1}{|\mathcal{N}(j)|}\sum_{i \in \mathcal{N}(j)} \mathbf{m}_{ij}^{(\ell)}.
\]

\paragraph{Equivariant layer normalization and gating.}
After aggregation, we normalize each angular-order block separately (suppressing the layer index $\ell$ for readability). Scalar channels ($l=0$) use channel-wise layer normalization, while higher-order channels ($l>0$) use root mean square (RMS) normalization without centering:
\[
\bar{\mathbf{m}}_{j,0} =
\gamma_0 \odot
\frac{\mathbf{m}_{j,0}-\mu_j}{\sqrt{\sigma_j^2+\varepsilon}}
+\beta_0,
\]
\[
\bar{\mathbf{m}}_{j,l} =
\gamma_l \odot
\frac{\mathbf{m}_{j,l}}{
\sqrt{\mathrm{RMS}\!\left(\mathbf{m}_{j,l}\right)^2+\varepsilon}},
\qquad l>0,
\]
where $\mu_j$ and $\sigma_j^2$ are the mean and variance across the scalar channels of node $j$, $\gamma_0$ and $\beta_0$ are learnable per-channel affine parameters, $\gamma_l$ is a learnable per-channel scale shared across all $2l+1$ components of each channel, $\odot$ denotes element-wise multiplication, and $\varepsilon$ is a small constant for numerical stability. At each node $j$, a separate RMS is computed for each angular order $l > 0$ by averaging the squared values over all channels of that order and their $2l+1$ components within node $j$, then taking the square root. Avoiding centering and applying a uniform scale across each irrep's components preserves directional structure and therefore equivariance. The normalized representation is then passed through an equivariant gated nonlinearity to yield the gated message $\tilde{\mathbf{m}}_j^{(\ell)}$.

\paragraph{Adaptive gated residual mixing.}
For hidden-to-hidden blocks where input and output irreps match, the model uses an adaptive gated residual connection. Gates are computed from scalar features of the current node state and the aggregated message, then broadcast across the tensor components of each irrep:
\[
\mathbf{h}_j^{(\ell+1)}
=
\mathbf{g}_j^{(\ell)} \odot W_{\mathrm{sc}}\mathbf{h}_j^{(\ell)}
+
\left(1-\mathbf{g}_j^{(\ell)}\right)\odot \tilde{\mathbf{m}}_j^{(\ell)},
\]
where $W_{\mathrm{sc}}$ is a learnable equivariant self-connection linear projection and $\mathbf{g}_j^{(\ell)}$ are the learned channel-wise gates. This lets the network decide, channel by channel, how much information to preserve from the previous representation versus incorporate from the local neighborhood.

\subsubsection{Absorber-conditioned pooling and invariant readout features}
After the final interaction block, the model constructs graph-level invariant features from absorber-local and context information. Let the final node embedding be decomposed into scalar, vector, and rank-2 tensor components,
\[
\mathbf{h}_i = \mathbf{s}_i \oplus \mathbf{v}_i \oplus \mathbf{t}_i.
\]
The absorber scalar state is averaged over absorber atoms,
\[
\mathbf{s}_g^{\mathrm{abs}} =
\frac{1}{|\mathcal{A}_g|}
\sum_{i \in \mathcal{A}_g}\mathbf{s}_i.
\]
The unconditioned initial absorber embedding is also averaged,
\[
\mathbf{z}_g^{\mathrm{abs},0} =
\frac{1}{|\mathcal{A}_g|}
\sum_{i \in \mathcal{A}_g}\mathbf{h}_{i,\mathrm{atom}}^{(0)}.
\]
For non-scalar channels, channel-wise Euclidean norms are computed over the $2l+1$ tensor components at each absorber site and averaged,
\[
\mathbf{n}_{g}^{(1)} =
\frac{1}{|\mathcal{A}_g|}
\sum_{i \in \mathcal{A}_g}\|\mathbf{v}_i\|,
\qquad
\mathbf{n}_{g}^{(2)} =
\frac{1}{|\mathcal{A}_g|}
\sum_{i \in \mathcal{A}_g}\|\mathbf{t}_i\|.
\]
To incorporate broader structural context, the averaged absorber scalar representation is used as a graph-level query for attention pooling. The query is concatenated with each atom's scalar features and passed through an MLP to obtain raw attention scores. Absorber atoms are masked out, and a softmax over the remaining sites' raw scores yields the attention weights. Then, a weighted sum of the non-absorber scalar features using these weights forms the context vector $\mathbf{s}_g^{\mathrm{ctx}}$. Note that when a graph contains multiple absorber sites, they contribute to a single query and yield one spectrum.

The spectral readout then predicts one spectrum per graph from the combined representation
\[
\mathbf{z}_g =
\left[
\mathbf{s}_g^{\mathrm{abs}}
\,\|\, \mathbf{z}_g^{\mathrm{abs},0}
\,\|\, \mathbf{e}_{c,g}
\,\|\, \mathbf{s}_g^{\mathrm{ctx}}
\,\|\, \mathbf{n}_g^{(1)}
\,\|\, \mathbf{n}_g^{(2)}
\right],
\]
where $\|$ denotes vector concatenation, with the core-level term omitted when core-level conditioning is disabled.

\subsubsection{Absorber-Z-routed spectral readout}
The spectral readout predicts one intensity value for each processed energy bin. The model uses an absorber-$Z$-routed dense mixture of readout experts. For each absorber atomic number $Z$, a lookup table stores a trainable vector $\boldsymbol{\theta}_Z \in \mathbb{R}^K$ containing one score (logit) for each of the $K$ experts. For graph $g$, the absorber atomic number $Z_g^{\mathrm{abs}}$ selects the corresponding row, and a softmax converts its scores into expert weights,
\[
\boldsymbol{\pi}_g
=
\mathrm{softmax}\!\left(\boldsymbol{\theta}_{Z_g^{\mathrm{abs}}}\right),
\]
and the predicted intensity vector is
\[
\hat{\mathbf{y}}_g
=
\sum_{k=1}^{K} \pi_{g,k}\, f_k(\mathbf{z}_g)
\in \mathbb{R}^{N_E},
\]
where $f_k$ is the $k$-th MLP expert and $N_E$ is the number of points on the model energy grid. Each expert has one hidden layer with 256 SiLU-activated units. The same routing weights apply at every energy bin and for all sites and core levels of a given element. Changes in line shape within an element enter through $\mathbf{z}_g$, which carries the absorber environment and core-level embeddings. In the optional native-grid mode, targets are interpolated onto the global model grid and masked outside each sample's energy range, and stored bin-coverage counts can mask bins outside the training coverage for that absorber element and core level at inference.

\subsubsection{Auxiliary absolute-energy scalar head}
When enabled, the model predicts an auxiliary scalar used to place spectra on an absolute energy axis. The configured target can be either the physical/reference edge energy $E_0$ or the stored absolute energy origin $E_{\mathrm{origin}}$. We pretrained the model jointly on Materials Project K-, L$_2$-, and L$_3$-edge spectra. Because the L-edge records lacked a separate $E_0$ label, we used the lowest sampled absolute energy of each spectrum, $E_{\mathrm{origin}}$, as the target for all three edges.

For graph $g$ with absorber atomic number $Z$ and core level $c$, let $T_g$ denote the reference value of the selected energy target in eV. The standardized target is
\[
\eta_g = \frac{T_g-\mu_{Z,c}}{\sigma_{Z,c}},
\]
where $\mu_{Z,c}$ is the mean reference energy for that $(Z,c)$ element--core-level pair, and $\sigma_{Z,c}$ is the corresponding standard deviation (replaced by $1$~eV if it falls below $10^{-6}$~eV).

The auxiliary MLP $f_E$ predicts the standardized energy from the graph representation $\mathbf{z}_g$,
\[
\hat{\eta}_g = f_E\!\left(\mathrm{stopgrad}(\mathbf{z}_g)\right).
\]
The stop-gradient operation confines updates from the auxiliary loss to $f_E$. At inference, the predicted energy in eV is
\[
\hat{T}_g = \mu_{Z,c} + \sigma_{Z,c}\,\hat{\eta}_g.
\]
Adding $\hat{T}_g$ to each point on the relative energy grid places the predicted spectrum on an absolute energy axis.

\subsection{Training objective}
Let $\delta_{g,n}=\hat{y}_{g,n}-y_{g,n}$ be the residual between the predicted and reference normalized intensities of graph $g$ at energy bin $n$. On the uniform grid used for the benchmarks, define the finite-difference operators
\[
\begin{aligned}
(D_h^0\delta_g)_n &= \delta_{g,n},\\
(D_h^1\delta_g)_n &= \frac{\delta_{g,n+1}-\delta_{g,n}}{h},\\
(D_h^2\delta_g)_n &= \frac{\delta_{g,n+2}-2\delta_{g,n+1}+\delta_{g,n}}{h^2}.
\end{aligned}
\]
For the training loss and the reported gradient and curvature MSEs, we use a fixed reference spacing $h_{\mathrm{ref}}\simeq0.87248\,\mathrm{eV}$. Fixing $h_{\mathrm{ref}}$ keeps the relative scale of the intensity, slope, and curvature terms consistent across different energy grids, allowing the same loss weights $\lambda_{\nabla}$ and $\lambda_{\nabla^2}$ to be reused without retuning. For each order $q\in\{0,1,2\}$, the mean squared error for one spectrum is
\[
\mathcal{L}_g^{(q)}=\frac{1}{N_E-q}
  \sum_{n=1}^{N_E-q}\big[(D_{h_{\mathrm{ref}}}^{q}\delta_g)_n\big]^2.
\]
Using the actual grid spacing $h_{\mathrm{grid}}\simeq0.556\,\mathrm{eV}$ yields the slope and curvature residuals on the physical energy axis, whereas $h_{\mathrm{ref}}$ rescales the derivative-error magnitudes by a constant factor while preserving all comparisons on the same grid.

We minimize the batch average of the per-spectrum objective
\[
\mathcal{L}_g=\mathcal{L}_g^{(0)}+\lambda_{\nabla}\mathcal{L}_g^{(1)}
+\lambda_{\nabla^2}\mathcal{L}_g^{(2)}+\lambda_T\left|\hat{\eta}_g-\eta_g\right|,
\]
where $\lambda_{\nabla}, \lambda_{\nabla^2}, \lambda_T \ge 0$ are loss weights. The first-difference term penalizes errors in local slope, including the steep rise at the absorption edge, while the second-difference term penalizes curvature mismatches and increases sensitivity to peak sharpness. The final term is the $L_1$ error in the standardized energy target, with $\lambda_T=0$ for the iron-oxide benchmark.

\subsection{Training details}
\subsubsection{Pretraining on Materials Project spectra}
We trained the baseline from random initialization on 537,203 Materials Project spectra. The model had 9.68 million parameters, with three message-passing layers, channel multiplicities $(128,40,20)$ for angular orders 0, 1, and 2, and eight spectral readout experts. We used AdamW with an initial learning rate of $10^{-3}$, weight decay of 0.01, and a global batch of 48 spectra (three per GPU across 16 GPUs). Training used bfloat16 mixed precision. We fixed the energy-origin loss weight at $\lambda_T=0.01$. 

The intensity, gradient, and curvature losses have different numerical scales, so their relative weights determine how strongly each term contributes to the objective~\cite{kendall2018multi}. To stabilize early training, we used a schedule inspired by curriculum learning~\cite{bengio2009curriculum} to fit the intensity profile before emphasizing local slope and curvature. We set both derivative weights to zero during a five-epoch warm-up, then increased them linearly over ten epochs, reaching $\lambda_{\nabla}=2$ and $\lambda_{\nabla^2}=1$ at epoch~16. From that epoch onwards, we began tracking an exponential moving average (EMA) of the model weights with decay 0.999. We used these EMA weights to evaluate the full objective on the 40,000-spectrum validation set after each epoch and halved the learning rate when the validation loss plateaued. The lowest validation loss under the full objective occurred at epoch~65. Training stopped at epoch~75 after ten consecutive epochs without improvement. We used the EMA weights from epoch~65 for single-precision (FP32) inference on the separate 40,000-spectrum test set.

\subsubsection{Transfer learning on iron-oxide spectra}
\label{sec:iron-oxide-transfer-methods}
For each FDMNES Fe K-edge spectrum, we constructed a graph of the corresponding iron-oxide structure and tagged the absorbing Fe site or sites. We kept all graphs from the same structure together when dividing the dataset into training, validation, and test sets. The held-out test set contained 861 spectra from 79 iron-oxide structures, including both bulk crystals and surface slabs. Of these structures, 74 contributed spectra for more than one individually tagged Fe site.

We used the same pretrained Materials Project checkpoint for zero-shot evaluation on the iron-oxide test set and to initialize both fine-tuning protocols. For readout-only adaptation, we froze the encoder and trained the spectral experts and their routing weights (1.13 million parameters) at an initial learning rate of $3\times10^{-4}$. For full fine-tuning, we made all 9.68 million parameters trainable and used an initial learning rate of $3\times10^{-5}$ to limit changes to the pretrained encoder. Using a lower learning rate for fine-tuning than for training from scratch is common in transfer learning~\cite{li2020rethinkingfinetuning} and has also been used for equivariant interatomic potentials~\cite{karimitari2026systematic}. After the first epoch of full fine-tuning, we resumed training on the same iron-oxide training set from the exponential moving average (EMA) of that run's weights.

We also trained two compact models from scratch at an initial learning rate of $10^{-3}$. Given the smaller size of the iron-oxide training set (4,169 spectra from 377 structures), we selected two compact architectures to limit the risk of overfitting. Both used three message-passing layers and an eight-expert readout. Their channel multiplicities were $(32,16,8)$ and $(64,20,10)$, giving 1.27 and 2.82 million parameters, respectively. All models trained on the iron-oxide data used AdamW, bfloat16 mixed precision, and the full gradient and curvature penalties from the first epoch. We omitted the energy-origin loss because the energy reference was fixed. We evaluated validation loss using EMA weights with a decay of 0.99, selected the checkpoint with the lowest validation loss, and used early stopping with a patience of 30 epochs.

\section{Results and discussion}
\subsection{Prediction across elements and absorption edges}
On the Materials Project test set, the direct-readout model predicted normalized intensities with a mean squared error (MSE) of $2.38\times10^{-3}$ and the absolute energy origin with a mean absolute error (MAE) of 0.14~eV. By comparison, predicting only the group mean $\mu_{Z,c}$ for each absorber element and core level yields an MAE of 0.76~eV on the same spectra; this 5.4-fold error reduction confirms that the auxiliary head captures local environment-dependent energy shifts rather than relying on the group average.

\begin{figure}[!htb]
    \centering
    \includegraphics[width=\linewidth]{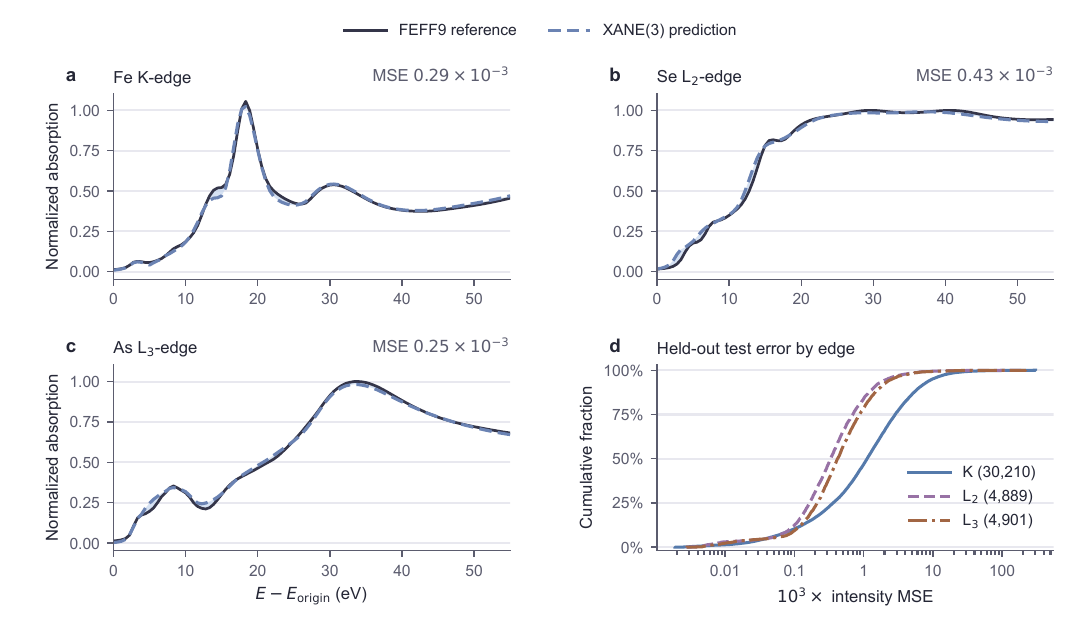}
    \caption{Reconstruction of held-out Materials Project FEFF9 spectra. (a--c) Fe K-, Se L$_2$-, and As L$_3$-edge spectra on the 100-point grid from 0 to 55~eV relative to the stored energy origin of each spectrum. Panels (a) and (c) show close peak alignment and intensity MSEs below their respective core-level medians in (d). The intensity MSE in (b) is above the L$_2$-edge median. (d) Empirical cumulative distributions of intensity MSE across all 40,000 held-out spectra, plotted by core level.}
    \label{fig:predictions}
\end{figure}

Figure~\ref{fig:predictions}a--c shows the Fe K-edge in CuFe$_2$O$_4$, the Se L$_2$-edge, and the As L$_3$-edge, respectively. Across the 40,000 held-out spectra, median intensity MSEs were $1.12$, $0.33$, and $0.42$ in units of $10^{-3}$ for K, L$_2$, and L$_3$ edges, respectively, as shown in Fig.~\ref{fig:predictions}d. The K-edge group spans 87 absorber elements, compared with 69 in each L-edge group, so these distributions reflect differences in absorber chemistry and environment as well as core level. The different line shapes in Fig.~\ref{fig:predictions}a--c also illustrate why the MSE alone gives an incomplete picture of how well individual spectral features are reproduced. Because the intensity MSE averages squared errors over all grid energy points, a large error at a narrow resonance can yield the same MSE as a smaller error spread across a broad absorption band.

The 18 elements represented only at the K-edge account for 64\% of the K-edge spectra and 78\% of the K-edge summed squared error. Across all 40,000 test spectra, with each spectrum weighted equally within its absorption edge, 70\% of the difference between the K-edge MSE and the average of the two L-edge MSEs falls in the first 20~eV above the stored energy origin. Restricting the comparison to K-edge spectra for the 69 elements represented at all three edges lowers the median MSE from $1.12\times10^{-3}$ to $0.29\times10^{-3}$. One hypothesis is that additional L$_2$- and L$_3$-edge training spectra for the same absorber element can improve K-edge prediction because the backbone weights are shared across edges. Before the first message-passing layer, XANE(3) projects the learned core-level embedding and adds it to the absorber node's scalar features, allowing the backbone to represent the same absorber site differently for distinct absorption edges. The readout concatenates the core-level embedding with its other inputs, while its expert routing weights depend only on the absorber element. L-edge training can therefore update both the shared backbone and the readout used for K-edge prediction.

Among the 69 absorber elements represented at all three absorption edges, the K-edge median MSE exceeds the L$_2$-edge median for 20 elements and the L$_3$-edge median for 19. Higher K-edge medians are more common among the 16 shared elements in period~4 of the periodic table, where the corresponding counts are 13 and 15. With all 69 shared elements weighted equally, the intensity MSE is highest at the K-edge ($1.87\times10^{-3}$ for K, $0.76\times10^{-3}$ for L$_2$, and $0.85\times10^{-3}$ for L$_3$). 

For these same 69 equally weighted elements, 99.5\% of the difference between the K-edge MSE and the average of the two L-edge MSEs falls in the 20--55~eV interval above the stored energy origin. Over the first 10~eV, the K-edge mean MSE is lower than the average of the two L-edge means. Restricting the analysis to the 60 elements with at least 20 spectra at each edge yields a similar result: the 20--55~eV interval accounts for 95\% of the gap. Within this subset, Mn and Fe together account for 73\% of the gap over the full 0--55~eV range. Across all 69 equally weighted elements, constant and slowly varying components of the spectral residuals account for 82\% of the K--L MSE gap. Most of the gap therefore reflects errors in the overall intensity level and the broad spectral shape rather than errors in fine spectral structure, such as sharp near-edge resonances.

\subsection{Ablation}
\label{sec:ablation}

Table~\ref{tab:ablation} compares the baseline and five ablations trained on the same Materials Project training split and evaluated on the same 40,000 held-out spectra. The intensity, gradient, and curvature MSEs measure spectral reconstruction on the relative energy grid. The MAE of the predicted energy origin, $E_{\mathrm{origin}}$, measures the error in placing that grid on the absolute energy axis.

\begin{table}[htbp]
\centering
\caption{Ablation study of XANE(3) on the Materials Project test set. The three spectral errors are reported as $10^3 \times \mathrm{MSE}$ and the energy-origin MAE is reported in eV. Bold values indicate the lowest error in each column.}
\begin{tabular}{@{}lcccc@{}}
\hline
Model & Intensity MSE & Gradient MSE & Curvature MSE & $E_{\mathrm{origin}}$ MAE (eV)\\
\hline
baseline & 2.38 & \textbf{1.19} & 3.31 & 0.14 \\
\hline
no derivative loss & \textbf{2.16} & 1.33 & 3.88 & 0.12 \\
no gated residual & 2.46 & 1.21 & 3.36 & 0.11 \\
no attention pooling & 2.45 & 1.20 & 3.33 & 0.13 \\
no layer norm & 3.37 & 1.57 & 4.14 & \textbf{0.10} \\
no tensorial features & 2.87 & 1.21 & \textbf{3.25} & 0.13 \\
\hline
\end{tabular}
\label{tab:ablation}
\end{table}
Across all six models, the energy-origin MAE ranged from 0.10 to 0.14~eV. The 0.04~eV spread was smaller than one-tenth of the 0.556~eV energy-grid spacing. All ablations therefore achieved comparable sub-eV accuracy in absolute energy placement despite clear differences in spectral reconstruction. Because the auxiliary head receives a detached graph representation and its loss does not update the backbone, these small variations in $E_{\mathrm{origin}}$ MAE do not reflect the primary spectral-reconstruction tradeoffs.

\begin{figure}[!htb]
\centering
\includegraphics[width=\linewidth]{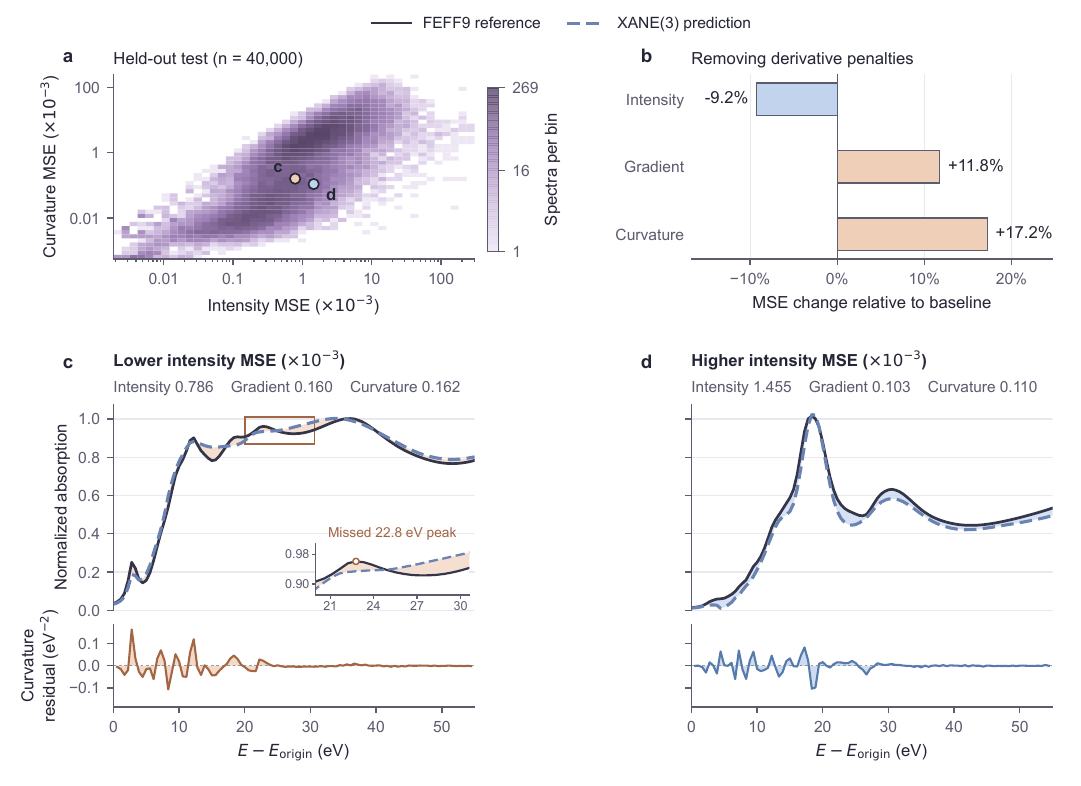}
\caption{Intensity and line-shape errors. (a) Joint distribution of intensity and curvature MSE across all 40,000 held-out Materials Project spectra. Color indicates the number of spectra per bin on a logarithmic scale. (b) Percentage changes in held-out intensity, gradient, and curvature MSE relative to the baseline when both derivative penalties are removed during training. (c,d) FEFF9 reference spectra and XANE(3) predictions for the two Fe K-edge examples marked in (a). Both predictions use the same baseline checkpoint trained with both derivative penalties. In (c), the FEFF9 spectrum has a local minimum at 15.0~eV, while XANE(3) predicts a shallower minimum at 15.6~eV. The inset in (c) highlights the local FEFF9 maximum at 22.8~eV that XANE(3) misses. The lower axes in (c,d) show prediction-minus-reference curvature residuals computed on the plotted energy grid, with identical vertical limits in both panels.}
\label{fig:mse_comp}
\end{figure}

\paragraph{Derivative-aware objective.} Removing the first- and second-derivative terms reduced the spectrum MSE from $2.38 \times 10^{-3}$ to $2.16 \times 10^{-3}$, a decrease of approximately $9.2\%$. However, the gradient MSE increased from $1.19 \times 10^{-3}$ to $1.33 \times 10^{-3}$, while the curvature MSE increased from $3.31 \times 10^{-3}$ to $3.88 \times 10^{-3}$, corresponding to increases of approximately $11.8\%$ and $17.2\%$, respectively, as shown in Fig.~\ref{fig:mse_comp}b. These results reveal a tradeoff between pointwise intensity accuracy and derivative-level agreement. Training exclusively on intensity MSE favors that metric, whereas including derivative penalties improves agreement in spectral slope and curvature at a modest cost in pointwise accuracy.

Intensity and curvature MSE generally rise together across the 40,000 held-out spectra (Spearman $\rho=0.80$, Fig.~\ref{fig:mse_comp}a). The two Fe K-edge predictions in Fig.~\ref{fig:mse_comp}c,d illustrate why the derivative terms are helpful: they provide complementary constraints on spectral shape since small pointwise intensity differences can coexist with noticeable differences in local slope or curvature. In panel c, the prediction has the lower intensity MSE but misses local maxima and minima between approximately 12 and 30~eV. These missed features are narrow and affect relatively few energy bins, so their contribution to the spectrum-wide intensity MSE is small, whereas they contribute disproportionately to slope and curvature errors (see Supplementary Information for the feature-width scaling). In panel d, a broad intensity deficit from 22 to 55~eV contributes to the intensity MSE over many bins, even where the predicted and reference curves have similar slopes and curvatures. On visual inspection, the prediction in panel d follows its reference line shape more closely than the prediction in panel c does, despite having a higher intensity MSE. Thus, evaluating intensity and derivative errors together gives a more complete picture of how well the model reproduces features such as peak positions, relative intensities, and post-edge oscillations.

To identify the absorber environments behind the larger errors in Fig.~\ref{fig:mse_comp}a, we traced the spectra back to their absorbing sites. O K-edge spectra account for 58\% of the spectra near the higher-error density maximum. Supplementary Fig.~1a illustrates how the model can locate the O K-edge rise yet round the first resonance. Such rounding can matter in $3d$ transition-metal oxides, where O K-edge features reflect O~$2p$--metal~$3d$ hybridization~\cite{de1989oxygen}. We also examined the 5\% of held-out spectra with the largest intensity MSE. Among Fe K-edge sites with oxygen-only first shells, sites with four O neighbors entered this high-error group more often than sites with six (Supplementary Fig.~1b). For Mg K-edges, 82 of 267 spectra fell in the high-error group. Supplementary Fig.~2 shows how the model can overestimate or underestimate post-peak absorption even for Mg sites with the same sixfold oxygen coordination.

\paragraph{Adaptive gated residual.} Starting with the second interaction block, the baseline uses a learned gate at each node to mix a projection of the previous node state with the processed neighborhood message, channel by channel. This allows each node to control how much new information to incorporate from its neighborhood into its current state. Successive layers extend the structural context available to each node by one graph hop at a time, similarly to how increasing the cluster radius in FDMNES brings more distant atoms into the XANES calculation. Thus, retaining site-specific features as this context expands may help limit oversmoothing in deeper networks. Replacing the adaptive gated residual connections with standard residual updates increased the spectrum MSE from $2.38 \times 10^{-3}$ to $2.46 \times 10^{-3}$. The gradient and curvature MSEs also increased slightly, from $1.19 \times 10^{-3}$ to $1.21 \times 10^{-3}$ and from $3.31 \times 10^{-3}$ to $3.36 \times 10^{-3}$, respectively. Adaptive gating therefore provides a modest improvement across the three spectral metrics in the reported experiment.

\paragraph{Attention pooling.} Absorber-query attention pooling uses the absorber representation to weight the non-absorber features that form its structural context. Replacing these learned weights with uniform weights increased all three spectral errors. In units of $10^{-3}$, the spectrum MSE rose from $2.38$ to $2.45$, the gradient MSE from $1.19$ to $1.20$, and the curvature MSE from $3.31$ to $3.33$.

Given the sensitivity of XANES to local coordination, we asked whether the learned weights preferentially emphasize sites closer to the absorber. Compared with uniform pooling, learned attention places slightly more cumulative weight on sites closer to the absorber, most noticeably at distances of 2--3~\AA{} (Supplementary Fig.~3a). To quantify this effect by edge, we averaged the absorber--site distances for each held-out graph twice, first weighting all non-absorber sites equally and then using their attention weights. Across the 30,203 eligible test K-edge spectra, the mean distance decreased from 5.10~\AA{} under uniform weighting to 4.92~\AA{} under attention weighting. The mean difference (attention minus uniform) was thus $-0.18$~\AA{} for K, while both L$_2$ and L$_3$ showed a small shift toward more distant sites (approximately $+0.075$~\AA{} in each group, Supplementary Fig.~3b). The K-edge site rankings also showed a modest preference for proximity. For each K-edge spectrum, we calculated the fraction of non-absorber sites at the minimum distance from the absorber. This fraction averaged 7.8\% across spectra and rose to 13.9\% when calculated among the highest-weight sites (Supplementary Fig.~3c). Thus, compared with uniform pooling, attention modestly favored shorter absorber--site distances for K-edge spectra and slightly longer distances for L$_2$- and L$_3$-edge spectra on average. Because the attention MLP receives only scalar node features rather than explicit absorber--site distances, this learned radial preference indicates that distance-dependent context weighting contributes to spectral accuracy.

\paragraph{Layer normalization.} Removing the custom equivariant layer normalization produced the largest deterioration across the three spectral metrics among the tested ablations. The spectrum MSE increased from $2.38 \times 10^{-3}$ to $3.37 \times 10^{-3}$, while the gradient and curvature MSEs increased from $1.19 \times 10^{-3}$ to $1.57 \times 10^{-3}$ and from $3.31 \times 10^{-3}$ to $4.14 \times 10^{-3}$, respectively. These changes correspond to increases of approximately $41.6\%$, $31.9\%$, and $25.1\%$. This result supports the importance of normalization for learning accurate spectral representations in the present architecture. The normalization scheme controls feature scales while preserving the transformation properties of scalar and higher-order channels, and its learnable scaling parameters allow their relative contributions to be adjusted during training. 

\paragraph{Tensorial versus scalar representations.} For this ablation, we removed the $l=1$ and $l=2$ channels and increased the scalar multiplicity to approximately preserve the total parameter count. The baseline used multiplicities $(m_0,m_1,m_2)=(128,40,20)$ and contained 9.68 million parameters, whereas the scalar-only model used $(198,0,0)$ and contained 9.64 million parameters. Matching parameter count accounts for the sharing of equivariant weights across the $2l+1$ spatial components of each irrep and enables comparison at approximately equal parameter budgets. Removing tensorial features increased the spectrum MSE from $2.38 \times 10^{-3}$ to $2.87 \times 10^{-3}$, an increase of approximately $20.6\%$. The gradient MSE increased slightly, from $1.19 \times 10^{-3}$ to $1.21 \times 10^{-3}$, whereas the curvature MSE decreased slightly, from $3.31 \times 10^{-3}$ to $3.25 \times 10^{-3}$. The clearest benefit of tensorial representations is therefore improved pointwise spectral reconstruction. Their effect on gradient agreement is comparatively small, and the scalar-only model achieves the lowest curvature MSE in the reported comparison.

Although the predicted spectrum is invariant under global SO(3) rotations of the full system, equivariant hidden features can encode directional relationships that contribute to invariant predictions through subsequent interactions and readout operations. Output invariance therefore does not make tensorial representations redundant. This improvement, however, carries a computational cost: adding tensorial features increased the measured time per epoch by approximately $46.7\%$ relative to the scalar-only model even though equivariant tensor-product operations were implemented using NVIDIA's cuEquivariance library with optimized CUDA kernels. 

\subsection{Transfer to iron-oxide surface and perturbed environments}
\label{sec:iron-oxide-transfer}
\begin{figure}[!htbp]
    \centering
    \includegraphics[width=\linewidth]{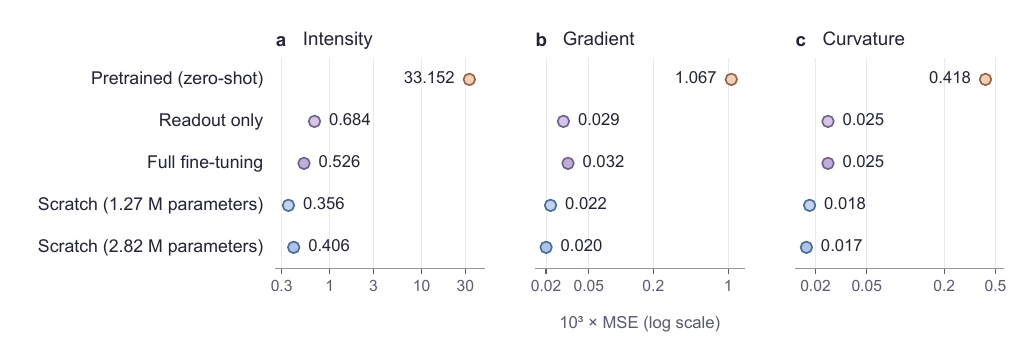}
    \caption{Transfer to 861 FDMNES Fe K-edge spectra from 79 held-out iron-oxide geometry groups. Panels show $10^3\times\mathrm{MSE}$ for (a) intensity, (b) gradient, and (c) curvature on logarithmic horizontal axes. The zero-shot model uses the Materials Project checkpoint. Readout-only and full fine-tuning both start from the same pretrained Materials Project checkpoint used for zero-shot evaluation. Readout-only fine-tuning freezes the encoder and trains 1.13 million readout parameters. In full fine-tuning, all 9.68 million parameters are trainable. For models trained from scratch, the labels give the total trainable parameter counts. All models are evaluated on the same target split, with one training seed per protocol.}
    \label{fig:iron-oxide-transfer}
\end{figure}

We tested whether an encoder pretrained on bulk crystals retains information about individual Fe sites on undercoordinated iron-oxide surfaces. Slabs provide a direct test of this question because terminating the parent crystal along a crystallographic plane leaves surface Fe atoms with fewer oxygen neighbors than in the bulk. The FDMNES target dataset spans bulk crystals, surface slabs, and configurations generated by atomic and lattice perturbations. We evaluated the Materials Project model under three transfer protocols (zero-shot prediction, readout-only adaptation, and full fine-tuning) and compared its performance with that of compact models trained from scratch, as summarized in Fig.~\ref{fig:iron-oxide-transfer}.

In zero-shot evaluation, the pretrained model yielded large errors across all three metrics, with an intensity MSE of $33.152\times10^{-3}$. Several factors explain this domain gap. Fe K-edge spectra accounted for less than 1\% of the Materials Project pretraining set, and the source and target spectra differ in how they were generated and aligned. For example, the first sampled energy of the source Fe K-edge spectra averaged 7114.37~eV, about 7.4~eV (13 grid intervals) above the fixed target origin of 7107~eV. Because a steep absorption rise shifted on the relative energy grid produces large pointwise residuals even for similar line shapes, this energy-origin offset, together with differences between FEFF9 and FDMNES and the order of normalization and interpolation, accounts for much of the zero-shot error. Zero-shot evaluation alone therefore does not separate the contributions of different energy alignments, simulation codes, preprocessing, and structural environments.

To test whether the bulk-pretrained encoder nonetheless retains transferable structural information, we froze the encoder and adapted only the spectral readout and routing weights (1.13 million of the 9.68 million parameters, or about 12\%). Relative to zero-shot prediction, readout-only adaptation reduced the intensity MSE by about 98\% to $0.684\times10^{-3}$, while gradient and curvature MSEs fell by factors of approximately 37 and 17, respectively. To verify that the frozen representation captured site-to-site differences rather than merely learning a global alignment shift, we evaluated a structure-averaged reference across the 74 of 79 held-out structures that contained separate FDMNES spectra for distinct, inequivalent Fe sites. For any predictor that cannot distinguish between inequivalent Fe sites within a structure, the lowest possible intensity MSE is achieved by assigning each structure its average reference spectrum. Even this optimal site-agnostic baseline yielded an intensity MSE of $4.97\times10^{-3}$ across the test set, approximately 7.3 times higher than the readout-only error. This shows that the frozen bulk-pretrained encoder preserves site-specific geometric information sufficient to distinguish inequivalent Fe absorbers within the same iron-oxide structure.

Unfreezing all 9.68 million parameters for full fine-tuning lowered the intensity MSE by a further 23\% (to $0.526\times10^{-3}$), while the gradient MSE increased by about 10\% and the curvature MSE remained unchanged at the reported precision. By contrast, both compact models trained from scratch outperformed either fine-tuning protocol across all three metrics: the 1.27-million-parameter model achieved the lowest intensity MSE ($0.356\times10^{-3}$), whereas the 2.82-million-parameter model achieved the lowest gradient and curvature MSEs. One likely reason for this advantage is that the compact models could specialize from the start of training on features governing Fe K-edge variations in iron oxides, such as oxygen coordination number and Fe--O bond lengths. During full fine-tuning, the pretrained model used an initial learning rate roughly 33 times lower than that used for training from scratch ($3\times10^{-5}$ versus $10^{-3}$), which may explain why full fine-tuning required 22 epochs to reach a validation loss below $2\times10^{-3}$, compared with 10 epochs for the smaller scratch model and 9 for the larger one.

Taken together, these results clarify the practical tradeoffs of bulk-to-surface transfer in XANE(3). Readout-only adaptation confirms that bulk-pretrained equivariant representations transfer across simulation codes and surface undercoordination well enough to resolve inequivalent absorber sites, which is useful for comparing candidate oxide structures when target data are scarce. When sufficient domain-specific spectra are available, however, compact models trained directly on the target dataset provide higher overall accuracy and faster convergence.

\section{Conclusions}
We have presented XANE(3), an $E(3)$-equivariant graph neural network that serves as an accurate and efficient surrogate for XANES simulation across diverse chemistries and absorption edges. On the Materials Project benchmark, incorporating higher-order tensorial features and equivariant normalization substantially improves pointwise spectral reconstruction, while the derivative-aware training objective enhances slope and curvature fidelity where narrow resonances carry key structural information. In transfer evaluations on an FDMNES Fe K-edge dataset spanning bulk and surface iron oxides, undercoordinated sites, and structurally perturbed configurations, readout-only adaptation confirms that bulk-pretrained representations retain transferable, site-specific information capable of distinguishing inequivalent absorbers. At the same time, compact models trained directly on the target domain achieve the lowest aggregate errors, clarifying both the utility and current limits of bulk-to-surface transfer and motivating systematic evaluation of transfer accuracy as a function of local coordination and structural distortion.

Because XANE(3) learns a mapping directly from atomic structure and composition, $f(\{\mathbf{r}, Z\})$, the same framework could be applied to other structure-dependent one-dimensional observables. In addition, the graph-based formulation naturally accommodates both periodic and non-periodic systems, making the approach applicable to crystals, surfaces, and molecular systems within a unified framework. Furthermore, XANE(3) could be integrated with downstream ML models for spectral interpretation and quantitative property inference, enabling end-to-end workflows that connect atomic structure, simulated spectra, and chemically meaningful descriptors. Such a framework could help accelerate forward spectral prediction, materials discovery, structure refinement, and data-driven spectroscopy more broadly.

\section*{Data availability}
The Materials Project FEFF9 spectra are available from Figshare for the K edge~\cite{mathew2018high} (\url{https://doi.org/10.6084/m9.figshare.5678998.v1}) and the L$_2$ and L$_3$ edges~\cite{chen2021database} (\url{https://doi.org/10.6084/m9.figshare.12824513.v1}). The iron-oxide benchmark contains 5,919 FDMNES Fe K-edge spectra from 539 geometries. The spectra, atomic structures, absorber-site indices, normalized targets, and training, validation, and test assignments are available at \url{https://huggingface.co/datasets/vitorgrizzi/xane3-iron-oxide}.

\section*{Code availability}
The XANE(3) source code for model training and evaluation is available at \url{https://github.com/vitorgrizzi/XANE3}. The pretrained Materials Project XANE(3) checkpoint used for the held-out FEFF9 benchmark and zero-shot iron-oxide predictions, together with inference scripts, is available at \url{https://huggingface.co/vitorgrizzi/xane3_mp}.

\section*{Acknowledgments}
We thank Drs. Massimiliano Delferro, David M. Kaphan and A. Jeremy Kropf for the fruitful discussions. This work was supported by the U.S. Department of Energy (DOE), Office of Science (SC), Office of Basic Energy Sciences (BES), Division of Chemical Sciences, Geosciences, and Biosciences (CSGB), Catalysis Science Program at Argonne National Laboratory under contract no. DE-AC02-06CH11357. L.N.P. acknowledges U.S. DOE Office of Science Graduate Student Research (SCGSR) award under contract number DE-SC0014664. The authors gratefully acknowledge the computing resources provided on Swing, a high-performance computing cluster operated by the Laboratory Computing Resource Center (LCRC) at Argonne National Laboratory and Argonne Leadership Computing Facility (ALCF).

\section*{Author contributions}
V.F.G. developed the model and software, performed the training, evaluation, and data analysis, prepared the figures, and wrote the original manuscript. L.N.P. generated the FDMNES iron-oxide dataset. J.X. advised on the research. C.L. supervised the project and contributed to the study design and manuscript revision.

\section*{Competing interests}
The authors declare no competing interests.

\bibliographystyle{unsrt}
\bibliography{refs}

\clearpage
\setcounter{figure}{0}
\setcounter{table}{0}
\setcounter{equation}{0}
\captionsetup[figure]{labelfont=bf}
\renewcommand{\figurename}{Supplementary Fig.}
\renewcommand{\thefigure}{\arabic{figure}}
\renewcommand{\theHfigure}{SuppFig.\arabic{figure}}

\section*{Supplementary Information}

\begin{figure}[htbp]
\centering
\includegraphics[width=\linewidth]{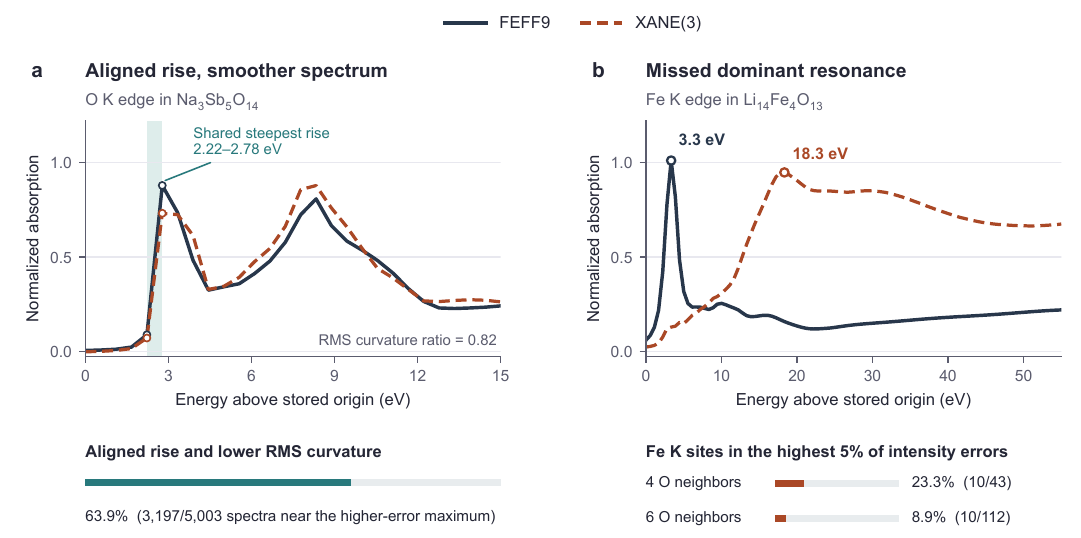}
\caption{\textbf{Early-rise alignment and dominant-resonance errors at the O and Fe K-edges.} Energies are relative to each spectrum's stored energy origin. \textbf{(a)} At the O site in Na$_3$Sb$_5$O$_{14}$, FEFF9 and XANE(3) have their steepest rise in the same interval (2.22--2.78~eV, shaded), whereas XANE(3) rounds the first resonance; its root-mean-square (RMS) curvature over 0--55~eV is 0.82 times the reference value. The bar summarizes the 5,003 spectra in the $5\times5$ bins centered on the higher-error density maximum in Fig.~3a of the main text, located by Gaussian smoothing of the counts on the log-binned grid ($\sigma=1.5$ bins). In 3,197 of these spectra (63.9\%), the predicted steepest rise falls within one grid interval (0.56~eV) of the reference while the prediction has lower RMS curvature. The steepest rise was located from unsmoothed forward differences in the first 15~eV, and curvature was computed from centered second differences on the full grid. \textbf{(b)} For the Fe site in Li$_{14}$Fe$_4$O$_{13}$, XANE(3) misses the dominant FEFF9 resonance at 3.3~eV and predicts a broad maximum at 18.3~eV (open circles). The bars show the fraction of Fe K-edge sites with four or six oxygen neighbors in an oxygen-only first shell whose intensity mean squared error (MSE) falls in the highest 5\% across all 40,000 held-out spectra. Periodic neighbors within 1.20 times the shortest absorber--neighbor distance are included in the first shell.}
\label{fig:sup_rise_resonance}
\end{figure}

\smallskip
\noindent{\small\textbf{O and Fe K-edge environments.} O K-edges account for 58\% of the spectra near the higher-error density maximum in Fig.~3a of the main text, where FEFF9 features are sharper than near the lower-error maximum. Across the full held-out set, O K-edges also account for 659 of the 2,000 spectra with the highest intensity MSE, the largest count for any absorber--edge group. Within the O K-edge group, 12.2\% of sites with H as the nearest neighbor fall in this high-error tail, compared with 2.2\% of sites with P as the nearest neighbor. This difference persists when distinct geometries receive equal weight and when sites are grouped by cell size and reference-spectrum sharpness. Among Fe K-edge sites with oxygen-only first shells, those with four O neighbors enter the high-error tail more frequently than those with six (Supplementary Fig.~\ref{fig:sup_rise_resonance}b).\par}

\smallskip
\noindent{\small\textbf{Scaling of spectral errors with feature width.} To illustrate why narrow resonances contribute disproportionately to derivative losses relative to intensity MSE (Fig.~3 of the main text), consider a well-resolved, localized residual $e(E)=A\,f((E-E_c)/w)$ of fixed shape $f$, amplitude $A$, center $E_c$, and width $w$. Setting $u=(E-E_c)/w$ gives $e^{(n)}(E)=A w^{-n}f^{(n)}(u)$ and $dE=w\,du$, so that $\int |e^{(n)}(E)|^2\,dE=A^2 w^{1-2n}\int |f^{(n)}(u)|^2\,du$. The integrated squared intensity ($n=0$), slope ($n=1$), and curvature ($n=2$) errors thus scale with feature width as $w$, $w^{-1}$, and $w^{-3}$, respectively.\par}

\begin{figure}[htbp]
\centering
\includegraphics[width=\linewidth]{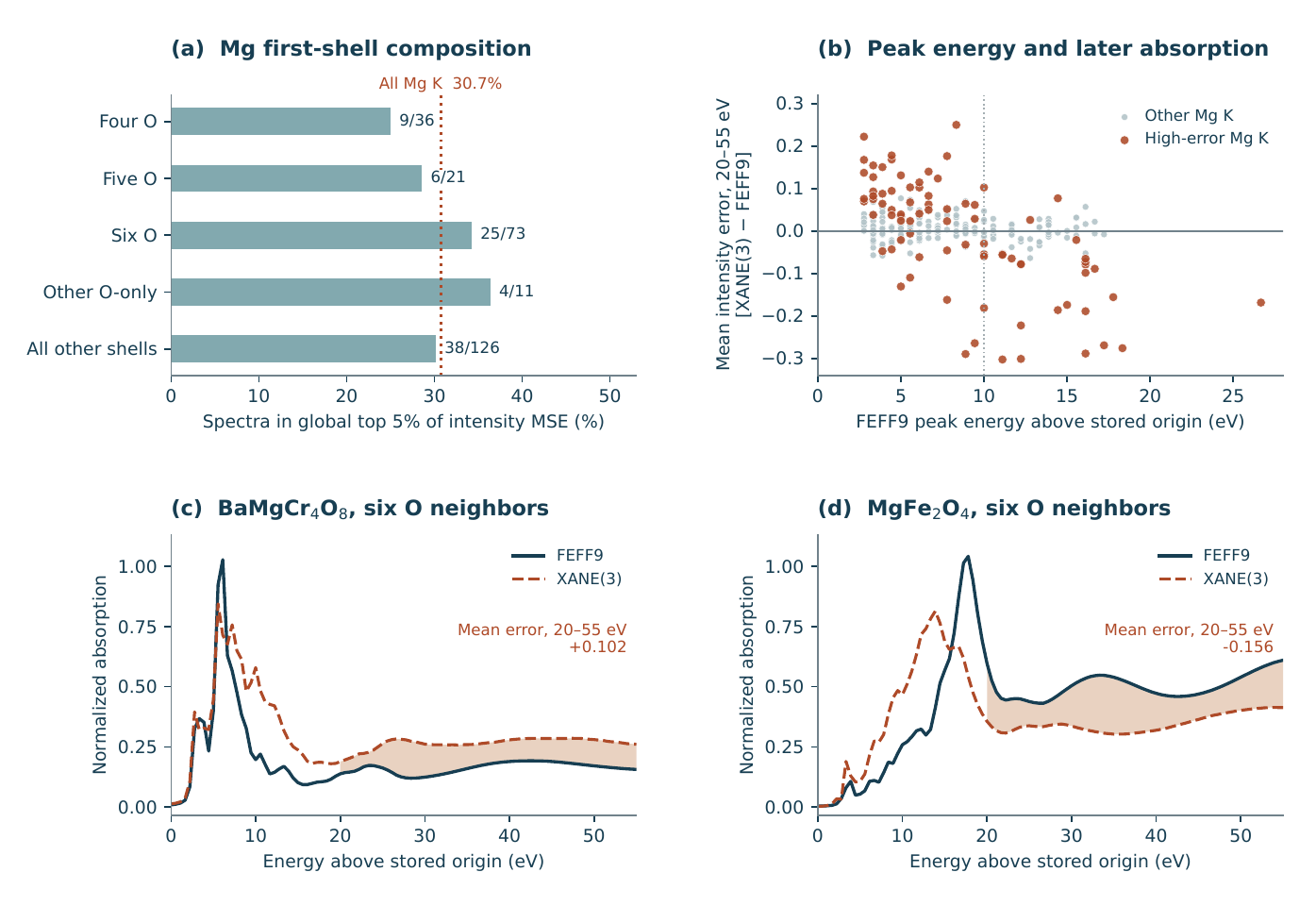}
\caption{\textbf{Mg K-edge errors across first-shell environments and reference line shapes.} \textbf{(a)} Of 267 held-out Mg K-edge spectra, 82 (30.7\%, dotted line) fall in the highest 5\% of intensity mean squared error (MSE) across the 40,000 held-out spectra. Each bar gives this fraction within one first-shell group, and labels next to each bar give the number of high-error sites out of all sites in that group. The first shell includes periodic neighbors within 1.20 times the shortest Mg--neighbor distance; the first four groups have oxygen-only shells. \textbf{(b)} Each point gives the energy of the FEFF9 maximum and the mean normalized intensity error (XANE(3) minus FEFF9) over 20--55~eV for one Mg K-edge spectrum, with energies relative to the stored energy origin. Among the 82 high-error spectra (orange), XANE(3) overpredicts mean absorption in 41 of 53 cases with a reference maximum below 10~eV and underpredicts it in 26 of 29 cases with a maximum at or above 10~eV. \textbf{(c,d)} Both Mg sites have six oxygen neighbors, and shading marks the difference between the curves over 20--55~eV. XANE(3) overestimates mean absorption in this interval for BaMgCr$_4$O$_8$ ($+0.102$) and underestimates it for MgFe$_2$O$_4$ ($-0.156$).}
\label{fig:sup_mg_k}
\end{figure}

\smallskip
\noindent{\small\textbf{Comparison across absorber--edge groups.} Among the 96 absorber--edge groups with at least 100 spectra, Mg K-edge had both the largest fraction in the full 40,000-spectrum test set's high-error tail (82/267) and the largest site-to-site variance in normalized FEFF9 intensity (computed at each of the 100 grid energies and then averaged). XANE(3) underestimates the contrast between rapid post-peak decay and sustained absorption, even among Mg sites with six O neighbors (Supplementary Fig.~\ref{fig:sup_mg_k}b--d). Evaluated within the 34,523-spectrum subset comprising these 96 groups, Mg K-edge ranked first by the fraction in the highest 5\% of raw intensity MSE (75/267); dividing each spectrum's MSE by its group's variance placed Mg K-edge sixth (30/267).\par}

\begin{figure}[htbp]
\centering
\includegraphics[width=\linewidth]{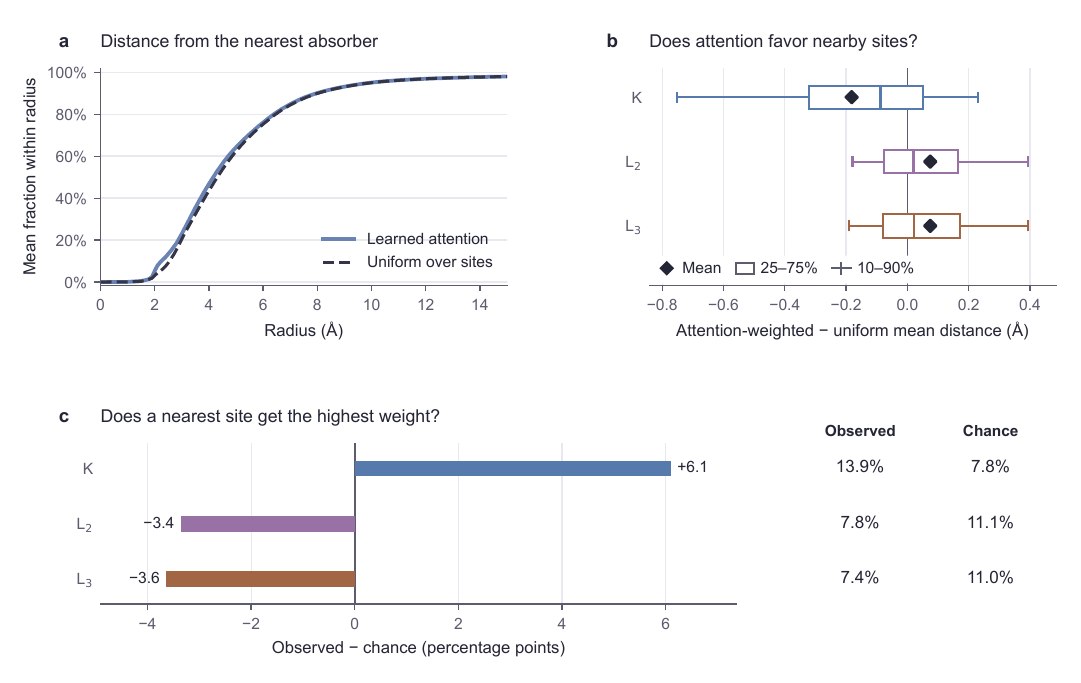}
\caption{\textbf{Distance dependence of absorber-query attention on the held-out Materials Project spectra.} \textbf{(a)} The solid curve gives the fraction of attention assigned to non-absorber sites within each radius of the absorber, averaged over the 39,987 spectra with at least one such site, while the dashed curve gives the fraction of non-absorber sites within the same radius. Distances are measured to the nearest periodic image of the absorber. \textbf{(b)} For each spectrum, we computed $\Delta r=\sum_i \alpha_i r_i-N^{-1}\sum_i r_i$, where $r_i$ is the distance of non-absorber site $i$ from the absorber, $\alpha_i$ is its attention weight, and $N$ is the number of non-absorber sites. Negative values indicate that attention assigns more weight to nearer sites on average. Boxes span the 25th to 75th percentiles, vertical lines within each box mark the median, whiskers span the 10th to 90th percentiles, and diamonds mark the mean. \textbf{(c)} Bars show how often a nearest site ranked first by attention weight relative to chance, expressed as the observed rate minus the chance rate in percentage points. For each spectrum, the observed rate is the fraction of highest-weight sites that are nearest, and the chance rate is the fraction of all non-absorber sites that are nearest. Sites within $10^{-4}$~\AA{} of the minimum distance count as nearest, and weights within $10^{-5}$ of the maximum on a relative scale count as tied. Each spectrum contributes equally within its edge group (30,203 K, 4,887 L$_2$, and 4,897 L$_3$ spectra).}
\label{fig:sup_pooling_attention_distance}
\end{figure}
\end{document}